\documentclass[10pt,twocolumn,letterpaper]{article}

\usepackage{iccv}
\usepackage{times}
\usepackage{epsfig}
\usepackage{graphicx}
\usepackage{amsmath}
\usepackage{amssymb}
\usepackage{booktabs}

\usepackage{overpic}
\usepackage{ftnxtra}

\usepackage[ruled,vlined]{algorithm2e}

\usepackage{amsmath,amsfonts,bm}

\def\eqref#1{equation~\ref{#1}}

\def\ceil#1{\lceil #1 \rceil}
\def\floor#1{\lfloor #1 \rfloor}
\def\1{\bm{1}}

\DeclareMathAlphabet{\mathsfit}{\encodingdefault}{\sfdefault}{m}{sl}
\SetMathAlphabet{\mathsfit}{bold}{\encodingdefault}{\sfdefault}{bx}{n}

\usepackage[pagebackref=true,breaklinks=true,letterpaper=true,colorlinks,bookmarks=false]{hyperref}

\iccvfinalcopy 

\def\iccvPaperID{6159} 

\begin{document}

\title{Sampling DETR: Exploring Spatial Sampling for Efficient Transformer Based Object Detectors}
\title{Sampling DETR: Spatially Adaptive Computation Allocation with Dynamic Feature Sampling}
\title{Sampling DETR: Efficient End-to-end Object Detection with Spatially Adaptive Sampling}
\title{Sampling DETR: Dynamic Computation Allocation with Spatially Adaptive Sampling}
\title{Sampling-DETR: Towards Dynamic Computation Allocation of Transformers}
\title{Poll-and-Pool Tokenization for Efficient Visual Analysis with Transformers}
\title{Poll-and-Pool Compression for Efficient End-to-End object Detection}
\title{Poll-and-Pool Sampling for Efficient End-to-End Object Detection}
\title{PnP-DETR: Towards Adaptive Computation Allocation of Transformers}
\title{PnP-DETR: Towards Efficient Visual Analysis with Transformers}

\author{\normalsize{Tao~Wang$^{1,3}$}\thanks{Work done during an internship at Yitu Tech.} \qquad \quad  \normalsize{Li~Yuan$^{4}$}\thanks{Corresponding author}  \quad  \qquad \normalsize{Yunpeng~Chen$^2$}  \quad  \qquad \normalsize{Jiashi~Feng$^4$} \quad  \qquad \normalsize{Shuicheng~Yan$^4$}\\
	\small{$^{1}$ Institute of Data Science, National University of Singapore}
	\small{$^{2}$ Yitu Technology} \\
	\small{$^{3}$ Integrative Science and Engineering Programme, NUS Graduate School, National University of Singapore} \\
	\small{$^{4}$ Department of Electrical and Computer Engineering, National University of Singapore} \\
	{\small \tt twangnh@gmail.com} \ \ {\small\tt ylustcnus@gmail.com} \ \ {\small\tt yunpeng.chen@yitu-inc.com} \\
	\ \ {\small\tt jshfeng@gmail.com} 
	\ \ {\small\tt shuicheng.yan@gmail.com}
}

\maketitle
\ificcvfinal\thispagestyle{empty}\fi

\begin{abstract}

	Recently, DETR~\cite{carion2020end} pioneered the solution of vision tasks with transformers, it directly translates the image feature map into the object detection result. 
	Though effective, translating the full feature map can be costly due to redundant computation on some area like the background.
	In this work, we encapsulate the idea of reducing spatial redundancy into a novel poll and pool (PnP) sampling module, with which we build an end-to-end PnP-DETR architecture that adaptively allocates its computation spatially to be more efficient.
	Concretely, the PnP module abstracts the image feature map into fine foreground object feature vectors and a small number of coarse background contextual feature vectors.
	The transformer models information interaction within the fine-coarse feature space and translates the features into the detection result. 
	Moreover, the PnP-augmented model can instantly achieve various desired trade-offs between performance and computation with a single model by varying the sampled feature length, without requiring to train multiple models as existing methods.
	Thus it offers greater flexibility for deployment in diverse scenarios with varying computation constraint. 
	We further validate the generalizability of the PnP module on \textbf{panoptic segmentation} and the recent transformer-based image recognition model {\textbf{ViT}}~\cite{dosovitskiy2020image} and show consistent efficiency gain. We believe our method makes a step for efficient visual analysis with transformers, wherein spatial redundancy is commonly observed.
	Code will be available at \url{https://github.com/twangnh/pnp-detr}.
\end{abstract}

\section{Introduction}


\begin{figure}[t]
	\centering
	\includegraphics[width=1\linewidth]{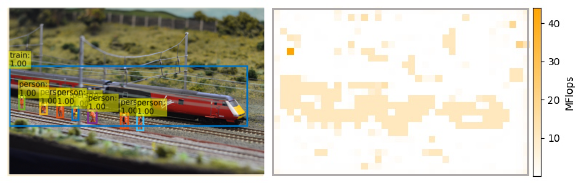}
	\caption{Left: Detection result. Right: Transformer computation density map. 
	Proposed method allows the model to adaptively allocate computation spatially and avoid computation expenditure on less informative background area.}
	\label{computation_density}
\end{figure}

Object detection is a fundamental computer vision task aiming to recognize object instances in the image and localize them with precise bounding boxes.
Modern detectors address this set prediction task mainly with proxy learning objectives, \textit{i.e.}, regressing offset from pre-defined anchor boxes~\cite{ren2015faster,lin2017focal} or boundaries from grid locations~\cite{tian2019fcos,zhou2019objects,duan2019centernet}.
Those heuristic designs not only complicate the model design but also require hand-crafted post-processing for duplicate removal. 
A recent method DETR~\cite{carion2020end} eliminates those hand-crafted designs and achieves end-to-end object detection.
It builds an effective set prediction framework on top of convolution feature maps with transformers~\cite{vaswani2017attention} and shows competitive performance to the two-stage Faster R-CNN~\cite{ren2015faster} detector.
The image feature map is flattened in the spatial dimension into one-dimensional feature vectors. 
The transformer then processes them with its strong attention mechanism to generate the final detection list.

Albeit simple and effective, applying the transformer networks to a image feature map can be computationally costly,
mainly due to the attention operation~\cite{vaswani2017attention} over the long flattened feature vectors.
These features may be redundant:  natural images often contain enormous background areas apart from the interested objects, which may occupy large part in the corresponding feature representation; also, some discriminative feature vectors may already suffice for detecting the objects.
Existing works improving the transformer efficiency mainly focus on accelerating the attention operation~\cite{kitaev2020reformer,katharopoulos2020transformers,wang2020linformer,choromanski2020rethinking}, and few of them consider the spatial redundancy discussed above. 

To address the above limitation, we develop a learnable poll and pool (PnP) sampling module. It aims to compress an image feature map into an abstracted feature set composed of fine feature vectors and a small number of coarse feature vectors. 
The fine feature vectors are deterministically sampled from the input feature map to capture the fine foreground information, which thus are crucial for detecting the objects.
The coarse feature vectors aggregate information from the background locations, and the resulting contextual information helps better recognize and localize the objects. 
A transformer then models the information interaction within the fine-coarse feature space and obtains the final result. 
As the abstracted set is much shorter than the directly flattened image feature map, the transformer computation is reduced significantly and mainly distributed over the foreground locations. 
Our approach is orthogonal to the approaches improving the transformer efficiency~\cite{kitaev2020reformer,katharopoulos2020transformers,wang2020linformer,choromanski2020rethinking} and can be further combined with them to obtain more efficient models.

Concretely, the PnP module is composed of two core sub-modules: a \textit{poll sampler} and a subsequent \textit{pool sampler}.
The \textit{poll sampler} incorporates a content-aware meta-scoring network that learns to predict the infromativeness score of the feature vector at each spatial location.
The feature vectors are then ranked spatially with the informativeness scores and a subset of most informative feature vectors are selected.
The subsequent \textit{pool sampler} dynamically predicts attention weights on the non-sampled feature vectors and aggregates them into a small number of feature vectors that summarize the background information.
Similar to the region proposal networks~\cite{ren2015faster},
the PnP module also aims to extract object-relevant information, but is end-to-end learned without explicit objective like object bounding box regression.
We build a PnP-DETR with the PnP module, which operates on the fine-coarse feature space and adaptively allocates its transformer computation in the spatial domain. 
Fig.~\ref{computation_density} is an example detection with computation density map (refer to Sec.~\ref{main_result_sec} for details of the map construction).
Existing methods of improving model efficiency still need train multiple models of different complexities for achieving various trade-offs of computation and performance.
Compared with them, the proposed PnP sampling allows the transformer to work with a variable number of input feature vectors and achieve instant computation and performance trade-off.

We conduct extensive experiments on the COCO benchmark, and the results show PnP-DETR effectively reduces the cost and achieves dynamic computation and performance trade-off. 
For example, without bells and whistels, a single PnP-DETR-DC5 obtains a 42.7 AP with 72\% reduction of transformer computation compared to the 43.3 AP baseline and competitive 43.1 AP with 56\% reduction.
We further validate the efficiency gain with panoptic segmentation and the recent vision transformer model (ViT~\cite{dosovitskiy2020image}). For example, PnP-ViT achieves near half of FLOPs reduction with only 0.3 drop of accuracy.  
To summarize, the contributions are:

\begin{itemize}

	\item We identify the spatial redundancy issue of the image feature map, which causes excessive computation of the transformer network in a DETR model. We therefore propose to abstract the feature map,
	so as to significantly reduce the model computation.
	
	\item To realize the feature abstraction, we design a novel two-step poll-and-pool sampling module. It first employs a poll sampler to extract the foreground fine feature vectors, and then utilizes a pool sampler to obtain the contextual coarse feature vectors. 
	
	\item We then build PnP-DETR, wherein the transformer operates on the abstract fine-coarse feature space and adaptively distributes the computation in the spatial domain. PnP-DETR is more efficient and achieves instant computation and performance trade-off with a single model, by varying length of the fine feature set.
	
	\item The PnP sampling module is general and end-to-end learned without explicit supervision like the region proposal networks~\cite{ren2015faster}. We further validate it on panoptic segmentation and recent ViT model~\cite{dosovitskiy2020image} and show consistent efficiency gain. We believe our method provides useful insights for future research on efficient solutions of vision tasks with transformers.

\end{itemize}

\section{Related Work}
\label{related_work}
\paragraph{Object Detection}

In recent years performance of object detection has been substantially improved~\cite{girshick2014rich,girshick2015fast,ren2015faster,liu2016ssd,lin2017focal,tian2019fcos,wang2019distilling,wang2019few} over traditional approaches~\cite{sung1998example,felzenszwalb2009object}.
Those modern methods mainly address the task with a relaxed learning objective, \textit{i.e.}, learning on a set of matched positive anchor box samples and predicting with post-processing (NMS) to suppress duplicates. The handcraft designs 
Recently, \cite{carion2020end} proposed an end-to-end DETR framework that learns an explicit set based objective with transformers~\cite{vaswani2017attention}, showing decent performance compared to previous two-stage methods~\cite{ren2015faster}. 
Our work aims at improving efficiency of end-to-end objectors by reducing spatial redundancy. Compared to most recent \textit{deformable} DETR~\cite{zhu2020deformable} that improves the attention efficiency, we aim to directly compress the feature map, which is from different perspective and could be potentially combined together. For example, by implementing bilinear interpolation kernel in the irregular sampled space~\cite{skorokhodov2020interpolating_git,skorokhodov2020interpolating} to enable the learning of deformable offset prediction.

\paragraph{Sparse Execution and Sampling}
Lots of works explored sparse execution in convolution layers~\cite{figurnov2016perforatedcnns,ren2018sbnet,cao2019seernet,xie2020spatially,figurnov2017spatially,dong2017more,figurnov2017spatially}, saving computation by avoiding convolution operations on some less informative spatial locations. In this work, we are partially inspired by the sparse convolution and explore sparse execution of transformers~\cite{vaswani2017attention} by developing a dynamic image feature sampling method for efficient subsequent processing.
Our work is also related to literature on learning a sampling policy for point cloud understanding tasks~\cite{dovrat2019learning,lang2020samplenet,nezhadarya2020adaptive}.
Different from these works where sampling is achieved by new data point generation, we directly address discrete sampling by using a novel sampling as ranking strategy.

\section{Method}
We first revisit the DETR~\cite{carion2020end}. Then we elaborate the proposed feature abstraction scheme, followed by detailed design of the PnP Sampling that realizes the abstraction.
Finally we illustrate the PnP-augmented models and their advantages. 
We denote constants, scalars, vectors, tensors and sets as upper-case, lower-case, bold lower-case, bold upper-case and blackboard-bold upper-case letters, respectively, \textit{e.g.}, N, i, $\textbf{f}$, $\textbf{F}$, $\mathbb{F}$.

\begin{figure*}[t]
	\centering
	\includegraphics[width=0.85\linewidth]{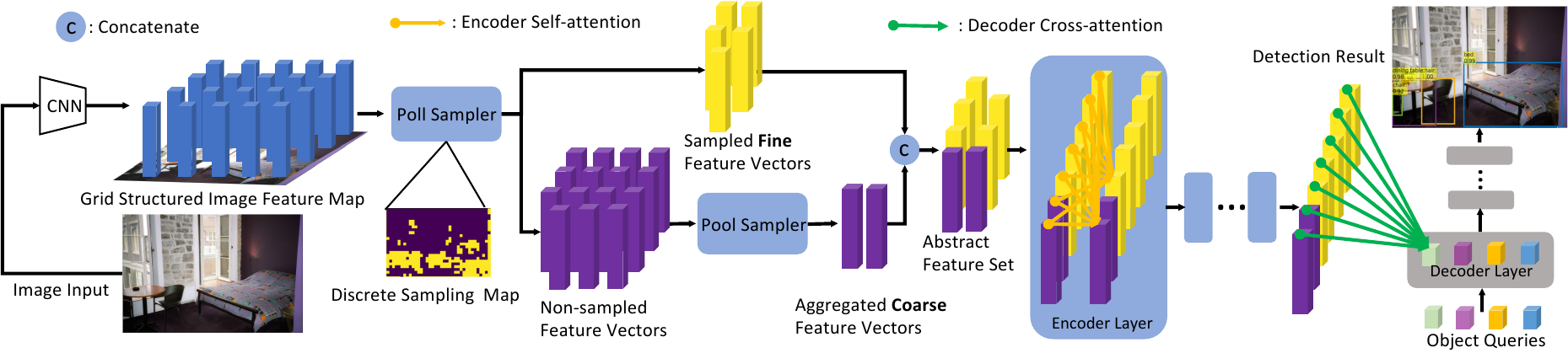}
	\caption{Illustration of the proposed PnP-DETR. The grid structured image feature map is first discretely sampled to obtain the fine feature vector set by a poll sampler, and the remaining non-sampled feature vectors are then aggregated into a small number of coarse feature vectors that summarize the contextual background information. The transformer encoder and decoder then operate on the fine-coarse feature space to model the information interaction and obtain the detection result.}
	\label{method_overview}
\end{figure*}

\subsection{Preliminaries} 
Without loss of generality, DETR~\cite{carion2020end} first utilizes a backbone convolution network $\mathcal{C}$ with parameters $\theta_c$ to extract the image feature map $\textbf{F}$:
\begin{equation}
\textbf{F}= \mathcal{C}(\textbf{I}, \theta_c)
\end{equation}
$\textbf{F}$ can be viewed as a grid structured feature vector set $\mathbb{F}$:
\begin{equation}
\mathbb{F}=\{\textbf{f}_{ij}\in\mathbb{R}^{C} |i=1,\ldots,H, j=1,\ldots,W\}
\end{equation}
Here  $\textbf{f}_{ij}$ is the feature vector at location $(i,j)$, $C$ is the number of feature channels,  $H$, $W$ are the height and width of the extracted image feature map. 
The grid-structured feature set $\mathbb{F}$ is then viewed as a set of high-level visual tokens with strong semantic information and translated into the detection result with a transformer $\mathcal{T}$ parametrized with $\theta_t$:
\begin{equation}
\{(cls_{k}, box_{k}) | k=1,\ldots,D\}= \mathcal{T}(\mathbb{F}, \theta_t)
\end{equation}
$(cls_{k}, box_{k})$ denotes one detected object with category and bounding box, the number of detections is fixed to $D$.

An intrinsic limitation of the grid structured visual token representation $\mathbb{F}$ is that it spans uniformly over the spatial locations and covers a large amount of background. 
Although the transformer can attend to different areas with its strong attention capability, the computation does not benefit from this advantage and is uniformly distributed over the spatial domain.
This deviates from our expectation that the processing power can be dynamically assigned to more relevant area like foreground locations while focusing less on area like background of a visual scene.

\subsection{Feature Abstraction} 
We propose a feature abstraction scheme to address the above limitation. 
It obtains two sets of feature vectors for compact feature representation:
\begin{equation}
\mathbb{F}_{f}=\{\textbf{f}_{n}\in\mathbb{R}^{C} | n=1,\ldots,N\}
\end{equation}
\begin{equation}
\mathbb{F}_{c}=\{\textbf{f}_{m}\in\mathbb{R}^{C} | m=1,\ldots,M\}
\end{equation}
The fine feature set $\mathbb{F}_{f}$ is discretely sampled from the full set $\mathbb{F}$, containing fine information that is essential for recognizing and detecting the objects.
The coarse feature set $\mathbb{F}_{c}$  is obtained by aggregating information from multiple spatial locations and encodes background contextual information.
Together, they form an abstraction set $\mathbb{F}^{*}$:
\begin{equation}
\mathbb{F}^{*}=\mathbb{F}_{f}\cup\mathbb{F}_{c}
\end{equation}
$\mathbb{F}^{*}$  encodes all necessary high-level information for detecting the objects within an image and is passed to a transformer for generating the object detection result.
Refer to supplementary for a theoretical analysis on the computation saving.
The feature abstraction scheme can also be viewed as a tokenization formulation that suits well for solving vision tasks with transformers.

\subsection{Poll and Pool (PnP) Sampling}
The above abstraction scheme need address two challenges.
\textbf{1)} The fine set requires deterministic binary sampling, which is non-differentiable. 
A handcrafted sampler can be learnt with some intermediate objectives, \textit{e.g.}, the region proposal networks~\cite{ren2015faster} or point proposal networks~\cite{zhou2019objects,duan2019centernet}, which is however incompatible with end-to-end learning, and the handcraft sampling rules may not be optimal. 
\textbf{2)} To extract a compact, coarse feature set only focusing on background contextual information is difficult. 
We divide the abstraction scheme into two steps and develop a poll sampler and a pool sampler to realize it.
The poll sampler first samples some feature vectors from the full set $\mathbb{F}$; the pool sampler then dynamically aggregates the remaining non-sampled feature vectors into a small number of coarse feature vectors. Fig.~\ref{method_overview} is an overview of the proposed method.
The samplers are deterministic and end-to-end learned with negligible computation cost.

\paragraph{Poll Sampler}
The poll sampler aims to obtain a fine feature set $\mathbb{F}_{f}$. 
Since explicitly learning a binary sampler is infeasible, we develop a sample as ranking strategy. 
We use a small meta-scoring network to predict the informativeness score for each spatial feature location $ (i,j)$:
\begin{equation}
s_{ij} =  \mathrm{ScoringNet}(\textbf{f}_{ij},\theta_{s})
\end{equation}
The larger the score is, the more informative the feature vector $\textbf{f}_{ij}$ is. We then sort all the scores $\{s_{ij}\}$ as 
\begin{equation}
[s_l, | l=1,\ldots,L], \aleph = \mathrm{Sort}(\{s_{ij}\})
\end{equation}
where $\aleph$ is the sorting order and $L=HW$. With $\aleph$, we then take the top $N$ scoring vectors to form the fine feature set:
\begin{equation}
\mathbb{F}_{f} = [\textbf{f}_l, | l=1,\ldots,N]
\end{equation}
To enable the learning of $\mathrm{ScoringNet}$ with back-propagation,  we take the predicted informativeness score as a modulating factor to the sampled fine feature set:
\begin{equation}
\mathbb{F}_{f} = [\textbf{f}_l*s_l, | l=1,\ldots,N]
\end{equation}
We find that normalizing the feature vectors before modulating can stabilize the learning of $\mathrm{ScoringNet}$:
\begin{equation}
\mathbb{F}_{f} = [LayerNorm(\textbf{f}_l)*s_l, | l=1,\ldots,N]
\end{equation}
We use layer normalization~\cite{ba2016layer} and turn off the affine parameters.
Ideally, $N$ may vary with the image content, but we observe that fixed amount sampling already generates good performance, \textit{i.e.}, $N=\alpha L$ where $\alpha$ is a constant fractional value, which we name as the poll ratio. 
This design also enables an extension to single model computation and performance trade-off discussed in Sec.~\ref{pnp_augmented}.

\paragraph{Pool Sampler}
The above poll sampler extracts the fine feature set. 
The remaining feature vectors mainly correspond to the background area. 
To compress them into a small feature set that summarizes the contextual information, we design a pool sampler that performs a weighted pooling of the remaining feature vectors to obtain a fixed number of $M$ background contextural feature vectors. 
This is partially inspired by the bilinear pooling~\cite{lin2015bilinear} and double attention~\cite{chen20182} operation where global descriptors are generated for capturing the second-order statistics of the feature map.
Formally, the remaining feature vector set is
\begin{equation}
\mathbb{F}_{r} = \mathbb{F} \setminus \mathbb{F}_{f} = \{\textbf{f}_r, | r=1,\ldots,L-N\}
\end{equation}
We project the feature vectors with a learnable weight $\textbf{W}^a\in \mathbb{R}^{C\times M} $ to obtain the aggregation weight $\textbf{a}_r\in \mathbb{R}^{M}$ :
\begin{equation}
\textbf{a}_r = \textbf{f}_r\textbf{W}^a
\end{equation}
and project the feature vectors with a learnable weight $\textbf{W}^v\in \mathbb{R}^{C\times C}$ to obtain the projected feature:
\begin{equation}
\textbf{f}^{'}_r = \textbf{f}_r\textbf{W}^v
\end{equation}
We then normalize the aggregation weight over all the remaining non-sampled locations with softmax:
\begin{equation}
a_{rm} =\frac{e^{a_{rm}}}{\sum_{r^{'}=1}^{N-L}e^{a_{r^{'}m}}}
\end{equation}
With the normalized aggregation weight, the projected feature vectors are aggregated to obtain a new feature vector that summarizes the information of non-sampled locations:
\begin{equation}
\textbf{f}_{m}=\sum_{r=1}^{L-N}\textbf{f}^{'}_r*a_{rm}
\end{equation}
By aggregating with all $M$ aggregation weights, we obtain the summarized coarse background contextual feature set:
\begin{equation}
\mathbb{F}_{c} = \{\textbf{f}_m, | r=1,\ldots,M\}
\end{equation}
It has been shown in~\cite{zhao2017pyramid} that the context information is crucial for recognizing the objects and is better aggregated by pyramid features of different scales.
Our pool sampler is able to freely obtain context information of different scales, by dynamically generating the aggregation weights. 
That is, some feature vectors may capture local context while others may encode global context.
We empirically show such an ability of the pool sampler by visualizing the aggregation weights.
Together with the fine set $\mathbb{F}_{f}$ from the poll sampler, the desired abstraction set $\mathbb{F}^*$  is obtained. Note that instead of convolution feature map, the PnP module can also be applied after a transformer layer.

\paragraph{Reverse Projection for Dense Prediction Tasks} The PnP module reduces the image feature map from 2D coordinate space to an abstracted space, which cannot be used for dense prediction tasks like image segmentation. To address the limitation, we propose to project the encoder output feature vectors back to the 2D coordinate space. Specifically, the fine feature vectors are scattered back to the sampled locations; the coarse feature vectors are first diffused back to original 2D space with the aggregation weight:
\begin{equation}
\hat{\textbf{f}}_{r}=\sum_{m=1}^{M}\hat{\textbf{f}_m}*a_{rm}
\end{equation}
and then scattered back the non-sampled locations of the poll sampler. $\hat{\textbf{f}_m}$ denotes output coarse feature vector from the encoder and $\hat{\textbf{f}}_{r}$ means the projected feature vector. The obtained 2D feature map is then used for dense prediction.


\subsection{PnP-augmented Models}
\label{pnp_augmented}
The PnP module is general and straightforward.
It can be plugged into existing models to enable them to operate on the fine-coarse feature space for better efficiency.
We here describe the models we build to evaluate the PnP module and our proposed random poll ratio scheme to enable instant computation and performance trade-off with a single model.

\paragraph{PnP-DETR and PnP-ViT} Recently~\cite{dosovitskiy2020image}\nocite{yuan2021tokens} introduced a transformer-based image recognition model named Vision Transformer (ViT).
We evaluate the generalizability of our method on the ViT model.
We build the PnP-DETR and PnP-ViT by plugging the PnP module before the transformer network.
The resulting models are end-to-end learned and other settings are the same with original models.
We use the hybrid ViT architecture~\cite{dosovitskiy2020image}.
Unlike original DETR and ViT wherein the transformer directly operates over the full image feature space, the PnP augmented transformer models the information interaction on the fine-coarse feature space and adaptively allocates its computation in the spatial domain to achieve better efficiency.

\begin{figure}[]
	\centering
	\includegraphics[width=0.95\linewidth]{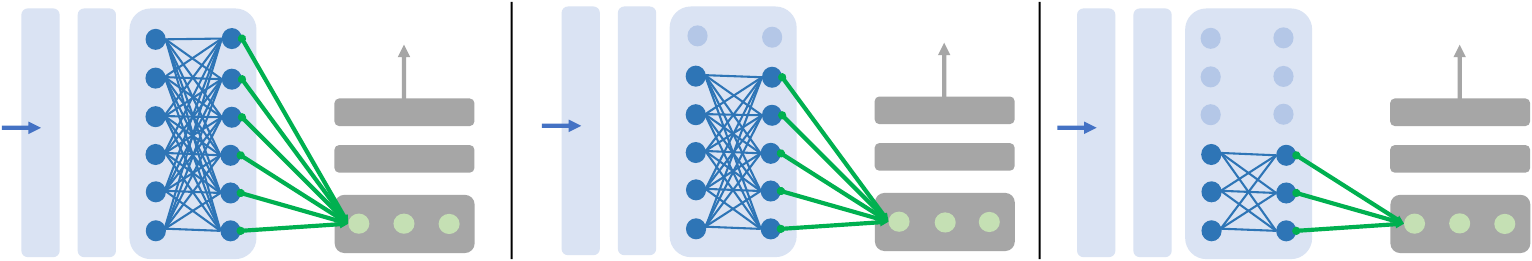}
	\caption{Instant computation-performance trade-off, by executing at different length. Blue: encoder layers Gray: decoder layers.}
	\vspace{-10pt}
	\label{instant_tradeoff}
\end{figure}

\paragraph{Instant Computation and Performance Trade-off} 
To achieve different computation and performance trade-offs, existing methods improving transformer efficiency generally train multiple models with different complexities controlling hyperparameters,
\textit{e.g.}, number of hashes in Reformer~\cite{kitaev2020reformer} and projected feature dimension in Linformer~\cite{wang2020linformer}. 
Unlike them, a model equipped with a PnP module can achieve instant single model computation and performance trade-off.
This is enabled by controlling the poll ratio $\alpha$ to determine the amount of fine information preserved.
With a larger $\alpha$, more fine feature vectors are obtained, and the overall performance is expected to be higher; with a smaller $\alpha$, the performance may be lower but more computation is saved.
However, we find inference with a different $\alpha$ to training severely degrades the performance. 
We propose to generate a random poll ratio during training: 
\begin{equation}
\alpha=uniform(\alpha_{low},\alpha_{high})
\end{equation}
Where $\alpha_{low}$ and $\alpha_{high}$ defines the value range. $\alpha$ is updated in each iteration. In this way, the transformer learns to work with variable length of input feature vectors, and thus achieves the desired single model computation and performance trade-off by inferring with different poll sample ratios (Fig.~\ref{instant_tradeoff}). The model only needs to be trained once.

\begin{table*}[h!]
	\centering
	\small
	\renewcommand{\tabcolsep}{5.0pt}
	\begin{tabular}{l|cccccc|cccc}
		\toprule
		Model & AP & AP$_{50}$ & AP$_{75}$ & AP$_{s}$ & AP$_{m}$ & AP$_{l}$ & F-encoder& F-decoder& F-sampler& F-total \\ \toprule
		DETR-R50~\cite{carion2020end} & 42.0 & 62.4  & 44.2  & 20.5  & 45.8  & 61.1  & 9.6G  & 1.9G  & -  &  11.5G  \\ 
		Deformable-DETR~\cite{zhu2020deformable} & 40.4 & 60.5 & 43.4 & 21.3 & 44.6 & 57.8 &  -  & - & -  & 5.5G (-52\%) \\
		PnP-DETR-R50-$\alpha$-0.33 & 41.1  & 61.5  & 43.7  & 20.8  & 44.6  & 60.0  & 3.2G  & 1.3G  & 0.1G  & 4.6G (\textbf{-60\%})   \\ 
		\quad \quad \small Inference-$\alpha$-0.5 & 36.1  & 59.8  & 36.1  & 13.9  & 38.7  & 57.7  &  -  & - & -  & - \\ 
		PnP-DETR-R50-$\alpha$-0.5 & 41.8  & 62.1  & 44.4  & 21.2  & 45.3  & 60.8  & 4.8G  & 1.5G  & 0.1G  & 6.4G (-45\%)   \\
		\midrule
		DETR-R50-DC5~\cite{carion2020end} & 43.3 & 63.1  & 45.9  & 22.5  & 47.3  & 61.1  & 69.2G  & 4.8G  & -  &  74.0G  \\
        ACT+MTKD(L=32)~\cite{zheng2020end} & 43.1 &  - &  - & 22.2  & 47.1  & 61.4 &  - &  - &  - &  58.2 (-21\%)  \\ 
        ACT+MTKD(L=24)~\cite{zheng2020end} & 42.3 &  - &  - & 21.3  & 46.4  & 61.0 &  - &  - &  - &  53.1 (-28\%)  \\
        Deformable-DETR-DC5~\cite{zhu2020deformable} & 42.1 & 62.3 & 45.6 & 24.3 & 45.6 & 57.3 &  -  & - & -  & 26.4G (-64\%) \\
		PnP-DETR-R50-DC5-$\alpha$-0.33 & 42.7  & 62.8  & 45.1  & 22.4  & 46.2  & 60.0  & 17.8G  & 2.5G  & 0.4G  & 20.7G (\textbf{-72\%})   \\ 
		PnP-DETR-R50-DC5-$\alpha$-0.5 & 43.1  & 63.4  & 45.3  & 22.7  & 46.5  & 61.1  & 29.1G  & 3.1G  & 0.7G  & 32.9G (-56\%)   \\

		\bottomrule
	\end{tabular}
	\caption{Results with fixed poll ratio training on COCO {\tt val} set. F-encoder, F-decoder, F-sampler, F-total denote the FLOPs of the encoder, decoder, PnP sampler and the full transformer, respectively. The FLOPs is obtained by averaging over the first 100 images of {\tt val} set. The backbone FLOPs is omitted as we focus on the transformer efficiency. Inference-$\alpha$-0.5 means inference with a mismatched poll ratio of 0.5 for PnP-DETR-R50-$\alpha$-0.33 model. \textit{Note we report single scale deformable DETR~\cite{zhu2020deformable} with 500 epochs training for fair comparison}, 
	the result is obtained with the official implementation. Refer to Sec.~\ref{related_work} for the relation between our method and deformable DETR.
	}
	\label{main_result}
\end{table*}


\begin{figure*}[t]
	\centering
	\begin{minipage}[c]{0.85\textwidth}
		\tiny
		\begin{overpic}[width=\textwidth]{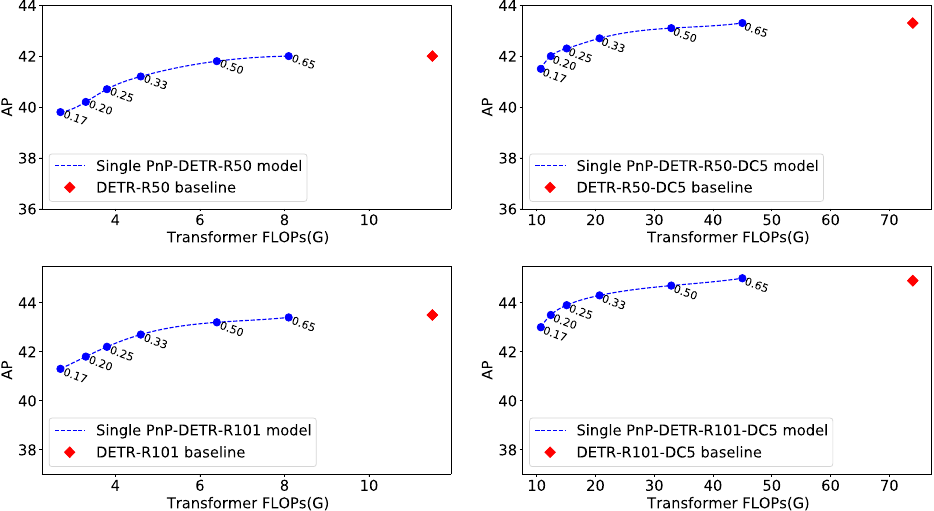}
			\put(16,41){
				\fontsize{6}{8}\selectfont
				\setlength\tabcolsep{0.5mm}
				\renewcommand{\arraystretch}{1}
				\begin{tabular}{cccccccc}
				\hline
        		- & baseline & 0.65 & 0.50 & 0.33 & 0.25 & 0.20 & 0.17\\
        		\hline
        		AP &42.0& 42.0 &  41.8 & 41.1 & 40.7 & 40.2 & 39.8\\
        		FLOPs (G) & 11.5& 8.1 &6.4 &4.6 &3.8 &3.3 & 2.7\\
				\hline
			\end{tabular}}
					\put(16,12.7){
			\fontsize{6}{8}\selectfont
			\setlength\tabcolsep{0.5mm}
			\renewcommand{\arraystretch}{1}
			\begin{tabular}{cccccccc}
				\hline
        		- & baseline & 0.65 & 0.50 & 0.33 & 0.25 & 0.20 & 0.17\\
        		\hline
        		AP &43.5& 43.4 &  43.2 & 42.7 & 42.2 & 41.8 & 41.3\\
        		FLOPs (G) & 11.5& 8.1 &6.4 &4.6 &3.8 &3.3 & 2.7\\
				\hline
		    \end{tabular}}
				\put(67.5,12.7){
    		\fontsize{6}{8}\selectfont
    		\setlength\tabcolsep{0.5mm}
    		\renewcommand{\arraystretch}{1}
    		\begin{tabular}{cccccccc}
    			\hline
        		- & baseline & 0.65 & 0.50 & 0.33 & 0.25 & 0.20 & 0.17\\
        		\hline
        		AP &44.9& 45.0 & 44.7 & 44.3 & 43.9 & 43.5 & 43.0\\
        		FLOPs (G) & 74.0& 45.0 &32.9 &20.7 &15.1 &12.4 &10.1\\    		
    			\hline
        	\end{tabular}}
        			\put(67.5,41){
        	\fontsize{6}{8}\selectfont
        	\setlength\tabcolsep{0.5mm}
        	\renewcommand{\arraystretch}{1}
        	\begin{tabular}{cccccccc}
        		\hline
        		- & baseline & 0.65 & 0.50 & 0.33 & 0.25 & 0.20 & 0.17\\
        		\hline
        		AP &43.3& 43.3 & 43.1 & 42.7 & 42.3 & 42.0 & 41.5\\
        		FLOPs (G) & 74.0& 45.0 &32.9 &20.7 &15.1 &12.4 &10.1\\
        		\hline
            \end{tabular}}
	    \end{overpic}
	\end{minipage}\hfill

	\begin{minipage}[c]{0.4\textwidth}
	\end{minipage}
\caption{Dynamic AP and FLOPs trade-off curve with single model trained with our method. The curve is obtained by evaluating with different poll ratios ($\alpha$) as denoted on the curve. The chosen $\alpha$ values roughly equals the fractions of $\frac{1}{6}$, $\frac{1}{5}$, $\frac{1}{4}$, $\frac{1}{3}$, $\frac{1}{2}$ and $\frac{1}{1.5}$.}
\vspace{-10pt}
\label{single_model_tradeoff}
\end{figure*}
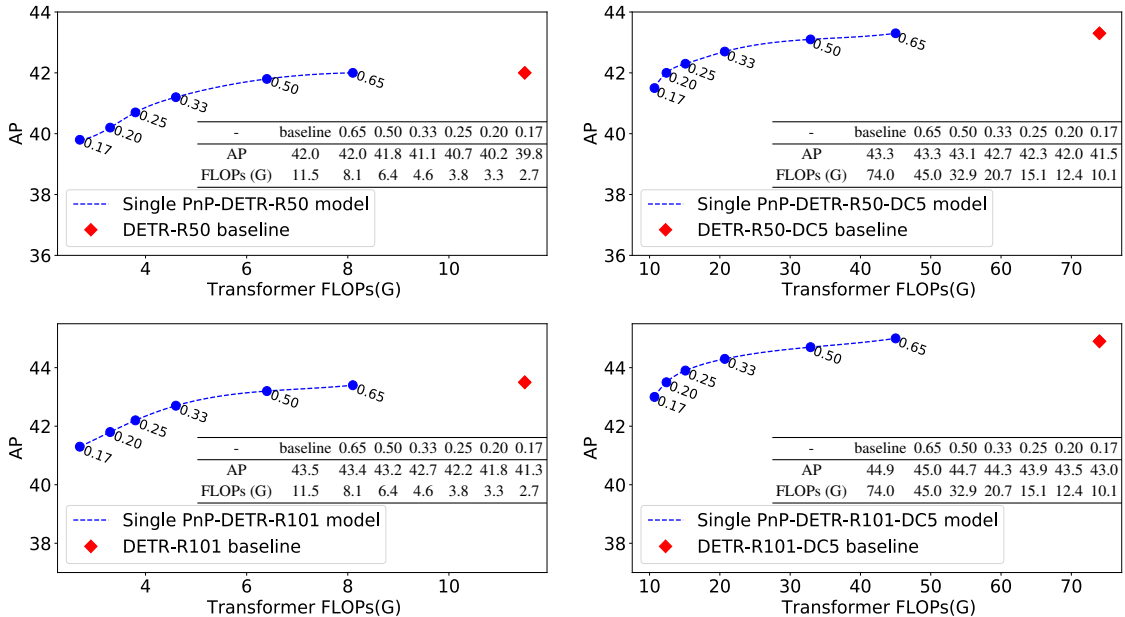

\section{Experiments}

\subsection{Implementation Details}
For training PnP-DETR, we use 4 images per GPU on 8-GPU machine, with a total batch size of 32. For training PnP-ViT, we use 32 images per GPU, with a total batch size of 256. The meta-scoring network is instantiated with a 2-layer MLP.
Unless otherwise stated, the pool sample number $M$ is set to 60 and 240 for R50 and R50-DC5 models, respectively.
Other settings including hyper-parameters, network architecture and loss functions follow the baselines for fair comparison.
Due to space limit, we defer more details like position embeddings to supplementary.


\subsection{Experiments on Object Detection}
\label{main_result_sec}

\paragraph{Fixed Poll Ratio Training}
Tab.~\ref{main_result} shows the results of the fixed poll ratio training on the COCO benchmark. For the DETR-R50 model, with an $\alpha$ = 0.33, PnP-DETR achieves 41.1 AP and 60\% reduction of transformer computation cost.
Further increasing $\alpha$ to 0.5, the performance reaches a similar level as the DETR baseline (AP of 41.8 vs. 42.0), with 45\% reduction of the computation. 
For DETR-R50-DC5 model, a similar trend is observed but more computation is saved.
We also evaluate the setting of mismatched training and test poll ratio.
The model trained with $\alpha$ = 0.33 gets nearly 5 AP drop when evaluating with $\alpha$ = 0.5.
This observation shows the necessity of applying random poll ratio training for the model to work with variable poll ratio. We also compare to the deformable DETR~\cite{zhu2020deformable}, as we did not incorporate multi-scale features, which is not the focus of this work, we compare to single scale deformable DETR for fair comparison.
Our method performs better than deformable DETR with less FLOPs, especially for large objects, \textit{e.g.}, AP$_l$ of 60.0 vs. 57.8 for the ResNet-50 backbone.


\paragraph{Dynamic Poll Ratio Training}
As shown in Fig.~\ref{single_model_tradeoff}, by training with the random poll ratio with a value range of $(0.15,0.8)$, the obtained model can achieve dynamic computation and performance trade-off by evaluating with variable poll ratio.
The AP for certain poll ratio is similar to the fixed poll ratio trained counterpart.
For example, a PnP-DETR-R50 model gets 41.1 AP with fixed poll ratio 0.33 training and 41.2 AP with random poll ratio training. 
The performance is the same to the baseline with a poll ratio of 0.65. 
We observe when the poll ratio is large, \textit{e.g.}, 0.5, increasing the poll ratio brings diminished gain in AP.
This is likely because the fine feature set already covers the essential spatial locations for detecting the objects, and thus more fine information only brings limited gain.
Similar observations are made with the ResNet-101 backbone. 
Tab.~\ref{tab:inference_time} shows the inference time compared to baseline model, the inference time is significantly reduced.

\begin{figure}[h]
	\centering
	\includegraphics[width=0.9\linewidth]{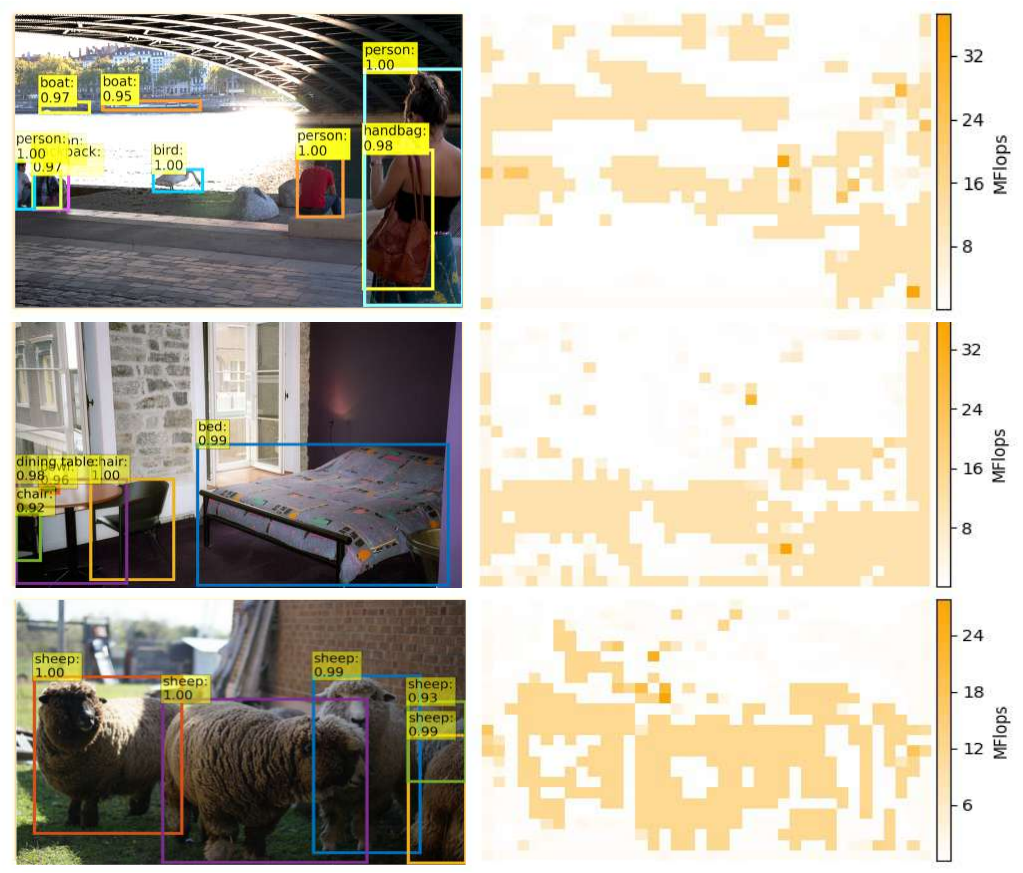}
	\caption{Example detection results and computation density maps with PnP-DETR-R50 model at poll ratio 0.33.}
	\label{detections_and_computation_density}
\end{figure}

\paragraph{Visualization of Computation Density Map}
Fig.~\ref{detections_and_computation_density} shows some example detection results and associated computation density maps, with poll ratio of 0.33. 
The objects are well detected while the computation is dynamically allocated to the spatial domain in a content-aware manner. 
To compute the density map, we assign a weight to each spatial location.
For poll sampled locations, the weight is 1. 
For each of other locations, the weight is the cumulative value of all pool sample aggregation weights at this location.
Then the transformer cost is distributed with the normalized weights to obtain the computation density map.

\subsection{Experiments on Other Tasks}
\paragraph{Panoptic Segmentation}
Following \cite{carion2020end}, we evaluate our method on the panoptic segmentation task.
To perform dense per-pixel segmentation as DETR, we project the encoder output feature back to the original 2D coordinate space. 
As shown in Tab.~\ref{panoptic}, the model saves computation and achieves instant performance and computation trade-off by varying the poll ratio $\alpha$, \textit{e.g.}, achieving Panoptic Quality (PQ) of 43.2 compared to 43.4 of a baseline DETR model, with 5G less FLOPs (\textit{i.e.}, 6.6G vs. 11.6G).

\paragraph{Image Recognition}
We also apply the PnP sampling to the recent transformer-based image classification model of ViT~\cite{dosovitskiy2020image}. 
We use the hybrid architecture with ResNet50-stage4 feature map (14x14) and train the model on the ImageNet-1k dataset from scratch.
We set the pool sample number to 10.
We train the PnP-ViT with random poll ratio in the value range of $[0.2,0.8]$.
As shown in Tab.~\ref{vit_result}, the PnP-ViT achieves dynamic computation and performance trade-off as observed with the DETR model. 
The results show the generalizability of PnP sampling design. 

\begin{table}[t!]
	\centering
	\small
	\renewcommand{\tabcolsep}{2.0pt}
	\begin{tabular}{lccc}
		\toprule
		Methods & Encoder & Decoder & PnP-sampler\\
		\midrule
        DETR (baseline)  & 72.4 & 11.1 & - \\
        PnP-DETR-$\alpha$-0.5  & \textbf{28.4} & 10.5 & 2.1 \\ 
        PnP-DETR-$\alpha$-0.33  & \textbf{17.4} & 10.3 & 2.0 \\ 
		\bottomrule	
	\end{tabular}
	\caption{Inference time (ms) measured on TITAN RTX GPU, with ResNet-50-DC5 backbone.}
	\label{tab:inference_time}
	\vspace{-10pt}
\end{table}

\subsection{Model Analysis}
We then provide several experimental analysis to better understand the proposed method. 
To save experiment time, we sample the COCO benchmark to obtain a smaller dataset and conduct all experiments on the \textbf{sampled COCO dataset}. We design a class-incremental sampling that helps preserve the data distribution. Due to space limit, we defer sampling details and more experiments to supplementary.

\begin{table}[]
	\renewcommand{\tabcolsep}{2.0pt}
    \small
	\centering
	\begin{tabular}{lcccccc}
		\toprule
		- & DETR & $\alpha$-0.65 & $\alpha$-0.5 & $\alpha$-0.33 & $\alpha$-0.25 & $\alpha$-0.2\\
		\midrule
		PQ &43.4&  43.5 & 43.2 & 42.8 & 42.4 & 41.8\\
		SQ &79.3&  79.2 & 79.1 & 78.9 & 78.7 & 78.4\\
		RQ &53.8&  53.8 & 53.4 & 53.0 & 52.4 & 51.7\\
		FLOPs (G) & 11.6& 8.3 & 6.6 &4.8 &4.0 &3.5\\
		\bottomrule
	\end{tabular}
	\caption{Results on panoptic segmentation. ResNet-50 backbone is used. $\alpha$-* means inferring with a variable poll ratio.}
	\label{panoptic}
\end{table}

\begin{table}[]
	\renewcommand{\tabcolsep}{3pt}
	\small
	\centering
	\begin{tabular}{lcccccc}
		\toprule
		- & ViT & $\alpha$-0.7 & $\alpha$-0.5 & $\alpha$-0.33 & $\alpha$-0.25 & $\alpha$-0.2\\
		\midrule
		Top1-Acc &82.2&  82.1 & 81.9 & 81.6 & 81.4 & 81.2\\
		FLOPs (G) & 10.0& 7.3 &5.5 &3.9 &3.2 &2.8\\
		\bottomrule
	\end{tabular}
	\caption{Results on ViT model with hybrid architecture based on ResNet-50. $\alpha$-* means a single PnP-ViT model with a variable poll ratio for inference.}
	\label{vit_result}
\end{table}

\begin{figure}[]
	\centering
	\includegraphics[width=0.8\linewidth]{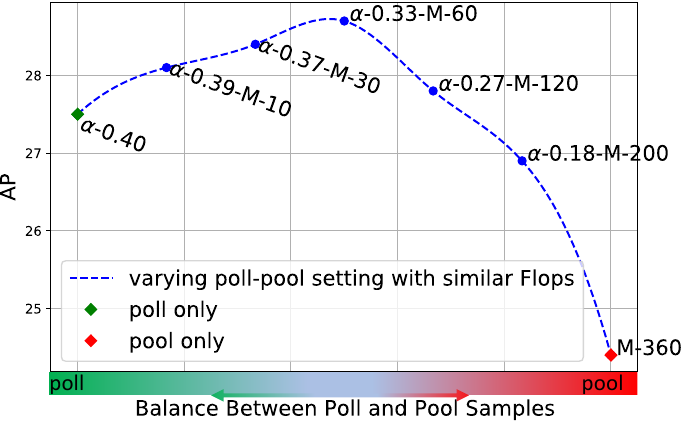}
	\caption{Varying the poll ratio ($\alpha$) and pool sample number ($M$) with the same amount of computation, with ResNet-50 backbone.}
	\label{poll_pool_balance}
	\vspace{-15pt}
\end{figure}

\begin{figure*}[t]
	\centering
	\includegraphics[width=0.90\linewidth]{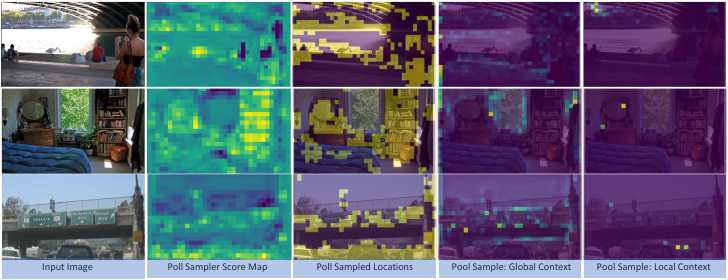}
	\caption{Visualization of poll sample locations and example aggregation weight map from the pool sampler, with PnP-DETR-R50.
	1st col: input images; 2nd/3rd cols: score maps of poll sampler and its sample maps correspondingly; last two columns: the example aggregation weight maps from the pool sampler, in which the former aggregates global context, while the latter aggregates local context.}
	\label{vis_poll_pool_samples}
	\vspace{-10pt}
\end{figure*}

\begin{figure}[]
	\centering
	\includegraphics[width=0.95\linewidth]{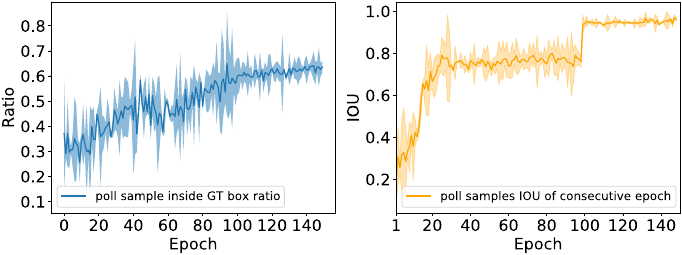}
	\caption{The learning dynamics of the poll sampler, with PnP-DETR-R50. The model is trained for 150 epochs with learning rate decay at 100 epochs. The left figure shows the proportion of sampled locations that lie within the GT bounding box areas. The right figure depicts the pixel IOU of sampled locations with previous epoch. The statistics are obtained on the \texttt{val} set.}
	\vspace{-10pt}
	\label{tracking_poll_sampler_learning}
\end{figure}

\paragraph{The Balance Between Poll and Pool Samplers}
As shown in Fig.~\ref{poll_pool_balance}, we vary the the poll sample ratio and the pool sample number to obtain the performance curve with the same amount of computation cost.
 We observe that \textbf{1)} with only poll sampling ($\alpha$-0.4), the performance is suboptimal; incorporating pool feature vector samples can significantly improve AP with the complementary background information from non-sampled locations, \textit{e.g.}, $\alpha$-0.39-$M$-10 model achieving about 0.7 AP higher than the $\alpha$-0.4 model.
\textbf{2)} with only pool sampling, the performance drops by a large margin. 
We assume it is difficult for the pool sampling to preserve accurate fine information, as it is designed to aggregate feature vectors spatially from different locations. 
\textbf{3)} the optimal setting is $1/3$ poll ratio with $60$ pool samples, indicating that a compact feature set should be mainly composed of fine feature vectors for accurate object detection.
We further individually examine the effects of pool sample number $M$ and poll sample ratio $\alpha$: \textbf{1)} We vary $M$ by fixing $\alpha$. \textbf{2)} We vary $\alpha$ by fixing $M$. Due to space limit, we defer the experiment results and analysis to the supplementary.

\paragraph{Visualizing Poll and Pool Sampling}
As shown in Fig.~\ref{vis_poll_pool_samples}, we visualize the poll sampler's scoring map, its sampled locations, and example aggregation weight map of the pool sampler.
To summarize, \textbf{1)} the poll sampler learns to sample the locations within and surrounding objects;
\textbf{2)} the pool sampler obtains different scales of context. For example, on the first row, the first pool sample attends to a wide range of spatial locations and encodes global context information; the second sample attends to a small area around the sky, and thus captures local context. 
We also have some other intriguing observations on the poll sampler: 
\textbf{1)} It learns to sample object alike area beyond the object categories used for training. For example, for the last row in Fig.~\ref{vis_poll_pool_samples}, locations around the traffic signs and the tree-like object are sampled.
The behavior is similar to a learned region proposal network (\textbf{RPN})~\cite{ren2015faster}, but learned without explicit supervision.
\textbf{2)} It tends to sample coarsely for some large and `easy' objects but finely for small ones. For example, fewer points are sampled for the woman in the first row and the bed in the second row; the books in the 
second image and the cars in the last image are smaller and more difficult to detect, so the poll sampler finely samples feature vectors for those objects and surrounding areas.

\paragraph{Tracking Poll Sampler Learning}
To better understand the learning process and dynamics of the poll sampler, we record two statistics during training: \textbf{(1)} the proportion of sampled locations that are within the GT bounding boxes;
\textbf{(2)} the pixel IOU of the sampled locations between consecutive epochs. 
As shown in Fig.~\ref{tracking_poll_sampler_learning}, we make following observations. 
\textbf{1)} The poll sampler gradually learns to sample more feature vectors that lie within the ground truth area but finally remains steady at about 60\%, indicating that it also attends some background and contextual locations that are crucial for recognizing and detecting the objects. 
\textbf{2)} The poll sampler initially has a large variation on its sampled locations, and thus the sampled areas of consecutive epochs have small IOU (\textit{i.e.}, about 0.2). 
During training, the IOU quickly converges to about 0.7 with around 30 epochs and remains steady at about 0.75, indicating that the sampler quickly learns to sample crucial feature vectors and the sampled locations does not change much. 
After learning rate decay at 100 epoch, the IOU of the consecutive 
epoch is close to 1.0, meaning the poll sampler converges.

\section{Conclusion}
In this paper, we encapsulate the idea of reducing spatial redundancy into a learnable PnP module. It is composed of a ranking based poll sampler that discretely samples fine feature information and a subsequent adaptive pool sampler that summarizes the background contextual information. The PnP module is general and can be incorporated into existing model for efficient processing while maintaining the performance, which is verified on object detection, panoptic segmentation and image recognition.
We believe the proposed method offers insights for future research into efficient visiual analysis with transformers.

{\small
\bibliographystyle{ieee_fullname}
\bibliography{egbib}
}

\clearpage

\section*{Computation Saving}

Here we show the concrete computation saving by the abstraction scheme, assume the length of the full feature set is $L=HW$ and the fraction of abstracted feature length is $r=(N+M)/L$. As shown in first row of Tab.~\ref{compute}, for encoder, since the complexity of self-attention layers is $\mathcal{O}(L^2)$ and the complexity of other layers (projection layers, feed-forward layers, normalization, \textit{e.t.c}) is $\mathcal{O}(L)$,we assume their actual computation cost is $aL^2$ and  $bL$ correspondingly. For the decoder, since the complexity of cross-attention is $\mathcal{O}(L^2)$, and the complexity of other parts is not related to the sequence length $L$, we assume their costs are $cL$ and a constant $O$ respectively.

Then with the abstracted feature set $\mathbb{F}^{*}$ as input, the computation cost of encoder self-attention is quadratically reduced to $ar^2L^2$, and the cost of other layers is reduced linearly to $brL$. For the decoder, the cross attention cost is reduced to $crL$, and the cost of other layers remains as $O$. The total computation of encoder compared to the original is 
\begin{equation}
\frac{ar^2L^2+brL}{aL^2+bL}=\frac{ar^2L+br}{aL+b} \in (r^2,r)
\end{equation}
With a larger sequence length $L$ the rate is more close to $r^2$ and more computation is saved. 

The total computation of decoder compared to original is 
\begin{equation}
\frac{crL+O}{cL+O}= \in (r,1)
\end{equation}
With a larger sequence length $L$ the rate is more close to $r$ and more computation is saved.

\begin{table}[]
	\centering
	\small
	\begin{tabular}{ccccc}
		\toprule
		& \multicolumn{2}{c}{encoder} & \multicolumn{2}{c}{decoder} \\ 
		\cmidrule(lr){2-3}\cmidrule(lr){4-5}
		input set	&  self-attn         &   o. layers       &   cross-attn        &   o. layers      \\ \toprule
		$\mathbb{F}$	&     $aL^2$      &     $bL$     &      $cL$      &      $O$    \\ 
		$\mathbb{F}^{*}$	&     $ar^2L^2$       &    $brL$      &    $crL$    &      $O$  \\\bottomrule 
	\end{tabular}
	\caption{Computation saving by the abstract feature set $\mathbb{F}^{*}$, compared to the full set $\mathbb{F}$. self-attn and cross-attn indicate the self-attention and cross-attention layers. o. layers denotes other layers except for the self-attention and cross-attention.}
	\label{compute}
\end{table}

\begin{algorithm*}[h!]
	\SetAlgoLined
	\textbf{Input}: 
	
	Cat2ImgID: mapping from category id to image id list (a dictionary)\;
	PerCatTHR: sampling threshold of image number (an integer)\;
	Cat2ImgIdSampled: empty mapping (a dictionary)\;
	SampledImgId: empty list (a list)\;
	SortedCatId: sorted category id on number of images, ascending (a list))\;
	$RandomSample$(Input,N): randomly sample input list to obtain subset with length N
	
	\textbf{Output}:
	
	SampledImgId: the sampled image ID list (a list)
	
	\For{Id in SortedCatId}{
		\If{Cat2ImgID[Id] $>$ PerCatTHR}{
			InSampled = [ImgId for ImgId in Cat2ImgID[Id] if ImgId in SampledImgId]\;
			NotInSampled = [ImgId for ImgId in Cat2ImgID[Id] if ImgId not in SampledImgId]\;
			
			\eIf{len(InSampled) $<$ PerCatTHR}{
				Cat2ImgIdSampled[Id]=InSampled+$RandomSample$(NotInSampled, PerCatTHR-len(InSampled))
			}{
				Cat2ImgIdSampled[Id]=InSampled
			}
		}
		
		SampledImgId+=Cat2ImgIdSampled[Id]
	}
	SampledImgId = set(SampledImgId)
	\caption{Class-Incremental Sampling Algorithm (pseudo code).}
	\label{sampling_alg}

\end{algorithm*}


\begin{figure*}[t]
	\centering
	\begin{minipage}[c]{0.95\textwidth}
		\begin{overpic}[width=\textwidth]{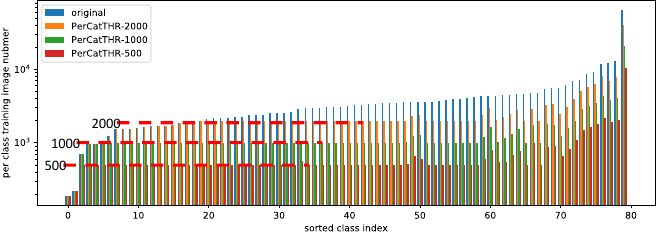}
			\put(25,31){
				\fontsize{8}{12}\selectfont
				\begin{tabular}{lcccccc}
		\hline
		-&original&PerCatTHR-2000&PerCatTHR-1000&PerCatTHR-500 \\
		\hline
		Total Image Num.&   ~118k    &   ~74k&   ~39k&   ~19k\\
		\hline	
			\end{tabular}}
	    \end{overpic}
	\end{minipage}\hfill

	\vspace{1pt}\begin{minipage}[c]{0.4\textwidth}
	\end{minipage}
\caption{Training image number distribution of the sampled COCO dataset obtained by the proposed class incremental sampling.}
	\vspace{10pt}
\label{incremental_sampling}
\end{figure*}

\section*{More Implementations}
Here we describe the implementation details about padding masks and position embedding.
For the fine feature set, we use the same sampling order of poll sampler to gather the corresponding position embeddings and padding masks. For the coarse feature set, we set the masks to $False$ to indicate that they are not paddings and employ pseudo position embedding by linearly combining position embeddings of the remaining feature set with the aggregation weight.

\subsection*{Class-Incremental Sampling on COCO Dataset}
\label{sampling_details}

In this section, we present the detailed about how we sample the COCO dataset to obtain a smaller version for faster experimental validation.
The COCO dataset has a skewed distribution of training image number over object categories, \textit{i.e.}, some categories have significantly smaller number of training images. Direct random sampling on all training images may cause too much loss of images on those scarce categories and the overall distribution may be even more biased. The mAP result on the biased dataset may be unstable and cannot well evaluate the model performance. 
To curcumvent the difficulty and obtain more effective sampled dataset, we design a new strategy. We rank the object categories according to their training image number, then perform an incremental sampling starting from the most scarce category to the most abundant category. The the sampling algorithm is given in Algorithm~\ref{sampling_alg}. Concretely, for each category, if the number of training images is more than a  sampling threshold number and the number of already sampled images for this category is less than the threshold number, then a sampling will be performed to obtain additional training images for reaching the threshold number. As shown in Fig.~\ref{incremental_sampling} is the distributions of obtained sampled versions of the COCO dataset, with different setting of the sampling threshold. The sampled dataset will be smaller given a smaller threshold. We use a sampling threshold of 500 to obtain a sampled COCO and conduct all the ablation experiments on the dataset. With the designed incremental sampling, the distribution of training images over most object categories is roughly uniform, and thus can be used to more stablly evaluate model performance than a randomly sampled sub-dataset while saving enormous experiment time.

\section*{Additional Ablations}

\paragraph{Pool Sample Number $M$ and Poll Sample Ratio $\alpha$}
To individually examine the effect of $M$ and $\alpha$, we conduct following experiments: \textbf{1)} varying $M$ by fixing $\alpha$. 
As shown in Tab.~\ref{num_pool}, compared to the model with only poll sample feature vectors ($M$-0), adding 30 pool feature vectors gets about 1 AP improvement, but when $M$ is larger than a certain value, the improvement is diminished (\textit{i.e.}, $60$). 
This phenomenon indicates that a small number of summarized feature vectors for the background contextual information is enough.
\textbf{2)} varying $\alpha$ by fixing $M$. As shown in Tab.~\ref{ratio_poll}, when the poll ratio $\alpha$ is small, increasing it significantly improves the performance (\textit{e.g.}, 25.2 AP to 27.1 AP by increasing $\alpha$ from 0.1 to 0.2).
This observation shows the importance of fine information for detecting the objects. 
When $\alpha$ is larger than about 0.5, the performance improvement is diminished, which is as expected since the feature vectors that rank lower mostly
correspond to the background locations, and thus the gain from including fine information on those locations is small.

\begin{table}[]
	\centering
	\small
	\renewcommand{\tabcolsep}{5.0pt}
	\begin{tabular}{lcccccc}
		\toprule
		-&AP&AP$_{50}$ &AP$_{75}$&AP$_s$ &AP$_m$&AP$_l$ \\
		\midrule
		baseline 			 &   29.1     &   48.0&   29.5&   10.9&   30.9&   44.2 \\
		$M$-0   &   27.3    &   46.7&   27.4&   9.4&   29.0&   42.9  \\
		$M$-30&   28.3    &   47.9&   28.7&   10.4&   29.9&   43.4  \\
		$M$-60&   28.7    &   48.4&   29.3&   10.5&   30.6&   44.4   \\
		$M$-120&   28.8    &   48.4&   29.2&   10.8&   30.4&   44.4 \\ 
		\bottomrule	
	\end{tabular}
	\caption{Effect of pool sample number ($M$), with ResNet-50 backbone. The poll sample ratio $\alpha$ is fixed at 0.33.
	}
	\vspace{-13pt}
	\label{num_pool}
\end{table}

\begin{table}[]
	\centering
	\small
	\renewcommand{\tabcolsep}{5.0pt}
	\begin{tabular}{lcccccc}
		\toprule
		-&AP&AP$_{50}$ &AP$_{75}$&AP$_s$ &AP$_m$&AP$_l$ \\
		\midrule
		baseline 			 &   29.1     &   48.0&   29.5&   10.9&   30.9&   44.2 \\
		$\alpha$-0.1   &  25.2    &   44.7&   24.2&   8.1&   26.0&   40.9  \\
		$\alpha$-0.2&    27.1     &   46.4&   27.3&   9.6&   29.0&   43.0  \\
		$\alpha$-0.3&   28.7    &   48.4&   29.3&   10.5&   30.6&   44.4   \\
		$\alpha$-0.5&   29.2     &   48.6&   29.2&   10.8&   30.3&   44.1 \\ 
		$\alpha$-0.7&   29.1 &   48.2&   29.3&   11.0&   30.9&   44.1 \\
		\bottomrule	
	\end{tabular}
	\caption{Effect of poll sample ratio ($\alpha$), with ResNet-50 backbone. The pool sample number $M$ is set as 60. The result is obtained with fixed poll ratio training.
		}
	\label{ratio_poll}
\end{table}

\paragraph{Different Architecture of Scoring Network}
As shown in Tab.~\ref{scorenet} is the result of different network architecture of the scoring network of the poll sampler, increasing the layer number from 1 to 2 improves the AP by 0.8 (\textit{i.e.}, 1-layer-fc and 2-layer-fc-256.). This is likely because the 2-layer network much more accurately predict the informativeness score. Further increasing the layer number gives diminished gain, \textit{i.e.}, 28.8 vs. 28.7 AP for 3-layer-fc-256 and 2-layer-fc-256. We also tried decreasing the hidden neuron unit number from 256 to 32, which reduces the computation, but the performance decreased, \textit{i.e.}, 28.2 for the 2-layer-fc-32 scoring network, which is 0.6 lower than the 2-layer-fc-256 network in AP. We choose the 2-layer-fc-256 network as the default architecture of the score network.

\begin{table}[t!]
	\centering
	\small
	\renewcommand{\tabcolsep}{3.0pt}
	\begin{tabular}{lcccccc}
		\toprule
		ScoreNet &AP&AP$_{50}$ &AP$_{75}$&AP$_s$ &AP$_m$&AP$_l$ \\
		\midrule
		1-layer-fc&   27.9    &   47.5&   28.3&   10.0&   29.6&   43.2   \\
		2-layer-fc-256&  28.7     &   48.4  &   29.3&   10.5 & 30.6 & 44.4 \\
		3-layer-fc-256&   28.8    &   48.4  &   29.4&   10.3 & 30.7 & 44.6 \\
		2-layer-fc-32&  28.2     &   48.0  &   29.1&   9.9 & 30.4 & 44.0 \\
		\bottomrule	
	\end{tabular}
	\caption{The effect of different scoring network architecture. For example, 1-layer-fc denotes 1-layer fully connected network (MLP), 2-layer-fc-256 means 2-layer fully connected network with 256 hidden neuron unit.}
	\label{scorenet}
\end{table}

\paragraph{Pool Sampler on The Full Feature Set}
While the proposed pool sampler operates on the non-sampled feature vectors of the poll sampler, it is interesting to see if directly applying the pool sampler on the full feature set for generating the coarse feature set would be better. As shown in Tab.~\ref{pool_on_full_set}, such setting leads to about 0.5 AP drop compared to the proposed two-step setting. This may be caused by the redundant information that have been captured by the fine feature vectors from polled samples.

\begin{table}[t!]
	\centering
	\small
	\renewcommand{\tabcolsep}{3.0pt}
	\begin{tabular}{lcccccc}
		\toprule
		-&AP&AP$_{50}$ &AP$_{75}$&AP$_s$ &AP$_m$&AP$_l$ \\
		\midrule
		pool-after-poll&   28.7    &   48.4&   29.3&   10.5&   30.6&   44.4   \\
		pool-full-set&   28.2    &   47.5&   28.7&   10.5&   29.9&   43.2  \\
		\bottomrule	
	\end{tabular}
	\caption{Result of applying pool sampler on the full feature set, compared to the proposed pool-after-poll design.}
	\label{pool_on_full_set}
\end{table}

\begin{table}[t!]
	\centering
	\small
	\renewcommand{\tabcolsep}{3.0pt}
	\begin{tabular}{ccccc}
		\toprule
		-&poll (proposed) & random & uniform & direct-interp\\
		\toprule
		w/o pool&27.3 & 22.9 & 25.9 & 26.1 \\
		w pool&28.7 & 23.9 & 26.2 & - \\
		\bottomrule	
	\end{tabular}
	\caption{Different alternative methods of the proposed poll sampling. The sampling ratio is set to 0.33 for all methods. w/o pool means removing the pool sampler. The random sampling result is obtained by an average of 3 runs.}
	\label{poll_alternatives}
\end{table}

\paragraph{Comparing the Proposed Sampling Strategies to Some Alternative Methods}
We compare the proposed poll sampler to some baseline alternatives including 1) random sampling: for each image, randomly sample the same amount of locations as the poll sampler and fix the sampled locations for training and evaluation. 2) uniform grid sampling: uniformly sample the 2D locations with equal interval. We adopt a general sampling mapping of $\floor{\frac{i}{\sqrt{r}}}\floor{\frac{j}{\sqrt{r}}}, i =0,1,...,\floor{W*\sqrt{r}},  j =0,1,...,\floor{H*\sqrt{r}}$ ($H$,$W$ are the height and width of the feature map and $r$ is the sampling ratio). With some specific poll ratio, the sampling is equavalent to MaxPooling, \textit{e.g.}, $r=1/4$ is equavalent to MaxPooling with kernel size 1 and stride 2. 3) direct interpolation: use interpolation to directly resize the feature map to target size $(\ceil{H/\sqrt{r}},\ceil{W/\sqrt{r}})$. 
As shown in Tab.\ref{poll_alternatives}, compared to proposed ranking based poll sampling, random sample leads to a large drop in AP, \textit{i.e.}, 22.9 vs 27.3 for the without pool sampling setting and 23.9 vs 28.7 for the with pool sampling setting. Uniform grid sampling and direct interpolation also generate lower performance than poll sampling, \textit{e.g.}, 25.9 and 26.1 compared to 27.3 under the without pool sample setting. The result shows the proposed poll sampler learns effective sampling policy and is better than those simple baselines.


\end{document}